\documentclass[10pt,twocolumn,letterpaper]{article}

\usepackage{cvpr}
\usepackage{times}
\usepackage{epsfig}
\usepackage{graphicx}
\usepackage{amsmath}
\usepackage{amssymb}
\usepackage{dsfont}
\usepackage{siunitx}
\usepackage{subcaption}
\usepackage{multirow}
\usepackage{caption}
\usepackage[breaklinks=true,bookmarks=false]{hyperref}
\graphicspath{{./images/}{./results}}

\def\1{\mathds{1}}

\def\eg{{e.g.$\;$}}

\cvprfinalcopy 

\newcommand{\myparagraph}[1]{\vspace{5pt}\noindent{\bf #1}}

\def\cvprPaperID{x}

\begin{document}

\title{Learning to Generate Images of Outdoor Scenes\\from Attributes and Semantic Layouts\\
}

\author{Levent Karacan$^1$ $\quad$ Zeynep Akata$^2$ $\quad$ Aykut Erdem$^1$ $\quad$ Erkut Erdem$^1$ \vspace{4mm} \\ 
$^1$Hacettepe University Computer Vision Lab, Beytepe Campus, Ankara, Turkey \\
$^2$Max-Planck Institute for Informatics, Saarland Informatics Campus, Saarbr\"ucken, Germany \vspace{2mm}\\
{\tt\small $^1$\{karacan,aykut,erkut\}@cs.hacettepe.edu.tr, $^2$akata@mpi-inf.mpg.de}
}
\maketitle

\begin{abstract}
Automatic image synthesis research has been rapidly growing with deep networks getting more and more expressive. In the last couple of years, we have observed images of digits, indoor scenes, birds, chairs, etc. being automatically generated. The expressive power of image generators have also been enhanced by introducing several forms of conditioning variables such as object names, sentences, bounding box and key-point locations. In this work, we propose a novel deep conditional generative adversarial network architecture that takes its strength from the semantic layout and scene attributes integrated as conditioning variables. We show that our architecture is able to generate realistic outdoor scene images under different conditions, e.g. day-night, sunny-foggy, with clear object boundaries.   
\end{abstract}

\section{Introduction}
\label{sec:intro}
\begingroup 
``\slshape{Maybe in our world there lives a happy little tree over there.}''\\
{\rightline{\rm --- {Bob Ross}}}\\
\endgroup

Automatically synthesizing realistic images has been an emerging research area in deep learning. Imagining an entire scene in the presence of discriminative properties of the scene such as ``sunny beach with mountains on the back'' is an ability that humans possess. As the most expressive artificial neural networks would presumably have human-like properties including imagination, automatic image generation research is a step towards this goal. Moreover, it is of practical interest as generated images would ideally augment data for various other tasks, e.g. image classification. 

Generating photo-realistic images of various object types has not yet been solved, however many successful attempts have been made. Generative Adversarial Nets (GANs)~\cite{GPMXWDOCB14} generated digits~\cite{GPMXWDOCB14}, faces~\cite{RMC16}, chairs~\cite{DSB15}, room interiors~\cite{RMC16} and videos~\cite{VPT16}. On the other hand, Variational Autoencoders (VAEs)~\cite{KW14} are combined with visual attention~\cite{GDGRW15} and have been extended to generating images based on textual descriptions~\cite{MPBS16}. Moreover, Pixel RNN~\cite{OKK16} has been proposed as an alternative model for the same. 

Deep neural networks take their strength from the availability of large image collections which stabilizes the learning. However, in some domains with limited number of images various complementary sources of information has been proposed to stabilize the learning. Recently, for the domain of fine-grained image generation, GAN conditioned on detailed sentences synthesizes realistic bird images~\cite{RAYLSL16} where visual training data was limited. Moreover, integrating textual GAN with bounding box and keypoint conditionals~\cite{RAMSSL16} allows drawing bird at the desired location. On the other hand, conditioining VAEs on discriminative object properties, i.e. attributes, has generated faces~\cite{YYSL16} with different hair color, beard or glasses, at different ages. 

\begin{figure}[t] 
\includegraphics[width=\linewidth]{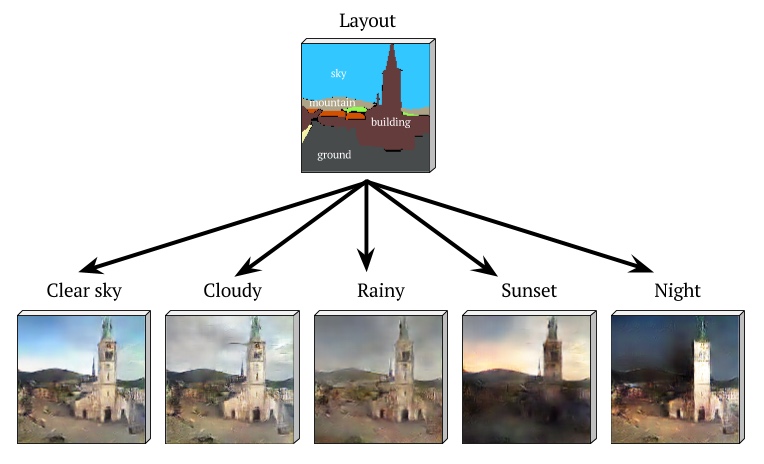}
\vspace{-8mm}
\caption{Our conditional generative adversarial network synthesizes realistic outdoor images from semantic layouts and transient scene attributes (Images generated automatically using a layout seen during training).}
\vspace{-3mm}
\label{fig:teaser}
\end{figure}

Apart from stabilizing the learning, conditioning variables also provide diversity to the generated images. Hence, we argue that the descriptive power of a generator network can be increased by conditioning it with respect to the object type, visual properties and location information. Object type conditioning teaches the network what to draw, visual properties specifies the visual details of the object and finally the location encodes where that object should be drawn. We propose a new GAN model architecture to generate realistic outdoor scenes, e.g. sea, mountain, urban scenes, conditioned on transient attributes, e.g. sunny, foggy, and on semantic layouts to determine the exact boundaries of where the object should be drawn. 
Our aim is to automatically generate outdoor scenes with various scene properties as shown in~\autoref{fig:teaser}. This problem has previously been tackled by designing hand-crafted procedures~\cite{LRTQH14}, however we propose to learn such transformations automatically through training deep convolutional networks. Towards this goal, we employ the recent ADE20K dataset~\cite{ZZPDBT16} that contains outdoor scenes with dense semantic layout annotations. To complement semantic layouts, we exploit a dataset of outdoor webcam sequences~\cite{LRTQH14} that provides per-scene attribute annotations. We complement the missing spatial layouts of~\cite{LRTQH14} with coarse semantic annotations of each scene and the missing attributes of~\cite{ZZPDBT16} with attribute predictions. We will make these supplementary annotations and our code publicly available.

Our contributions are summarized as follows. We propose a new conditioned GAN model that learns the content, i.e. transient attributes, to be drawn inside a scene layout. We show that our model generates realistic images of scenes with objects drawn within their own segments as well as transforming the scene by, for instance, imagining how a day scene would look like in the night.

\section{Related Work}
\label{sec:related}
We summarize published works on image generation and outdoor scene manipulation that are related to ours.

\myparagraph{Image Generation.} Since the generalization and expressive power of deep convolutional neural networks have been validated in various applications from image classification, detection, segmentation, research interests have expanded towards other challenging applications such as image generation. Several frameworks have been proposed to synthesize images from scratch. 

A convolutional image generator is proposed in \cite{DSB15} that maximizes the Euclidean distance between the real and generated 2D projections of chairs conditioned on type, viewpoint, etc. 
The first generation generative adversarial network (GAN)~\cite{GPMXWDOCB14} architecture has been designed as a two-player min-max game where a convolutional network learns to generate as realistic images as possible and a convolutional discriminator network learns to determine if an image is real or fake. Recently, different flavors of GANs have been proposed. In~\cite{ZKSE16}, the authors utilize image manifolds learned by GANs to define smoothness contraints. While they generate images through mouse strokes, we use semantic layouts. 3D-GAN~\cite{WZXFT16} extends GANs to 3D domain to generate 3D object silhouettes. CoGAN~\cite{LT16} extends GANs to learn a joint distribution of multi-domain images, e.g. learns a joint distribution of color and depth images, and a joint distribution of face images with different attributes. Moreover, GAN has been conditioned on different types of data sources. While DCGAN~\cite{RMC16} conditions GAN with class names, GAN-CLS~\cite{RAYLSL16} uses detailed natural language descriptions and GAWWN~\cite{RAMSSL16} uses bounding boxes and object keypoints. VGAN~\cite{VPT16} extends GANs to videos by conditioning the generation of future frames to the current frame. \mbox{S$^2$-GAN}~\cite{WG16} factorizes the image generation process into style and structure components. It combines two GANs, one for generating structure information, i.e. surface normals, and one for generating style information, i.e, appearance, where surface normals are used as a condition vector to generate indoor scenes. 

As an alternative to GANs, variational autoencoders (VAEs)~\cite{KW14} generate an image using a feed-forward convolutional decoder network and during inference the input is passed through the encoder that produces an approximate posterior distribution over the latent variables. The goal is to minimize the Euclidean distance between generated image and the posterior distribution. DRAW~\cite{GDGRW15} architecture combines a pair of recurrent neural networks with the VAE model for reading and writing portions of the image canvas at each time step. Given a single input image, DC-IGN~\cite{KWKT15} generates new images of the same object with variations in pose and lighting and disCVAE~\cite{YYSL16} conditions the image generation process with facial attributes. Finally, Pixel CNN~\cite{OKVEGK16} and Pixel RNN~\cite{OKK16} propose to generate image pixels sequentially.

\myparagraph{Outdoor Scene Editing.} 
As a high level image editing tool, in~\cite{LRTQH14}, the authors propose to train regressors that can predict the presence of attributes in novel images and develop a method that allows users to manually increase and decrease the effect of transient attributes of several outdoor scenes. They also introduce the Transient Attributes dataset, which includes images collected from the webcams viewing 101 scenes. As an alternative to regressors, a deep convolutional neural network is used in~\cite{BZGWJ16} to predict the transient attributes of an outdoor scene. In~\cite{LHERWC07}, a framework is presented for insertin new objects such as pedestrians into existing photographs of street scenes. Several outdoor scene datasets such as~\cite{ZZPDBT16, CORREBFRS16} may provide data to facilitate outdoor scene editing task. Cityscapes dataset~\cite{CORREBFRS16} is limited to street scenes, collected mostly to facilitate research on self driving cars. In our study, we decided to use ADE20K dataset~\cite{ZZPDBT16}, which provides dense segmentation of objects from indoor and outdoor scenes. 

\begin{figure*}[!t]
\centering
\includegraphics[width=\linewidth]{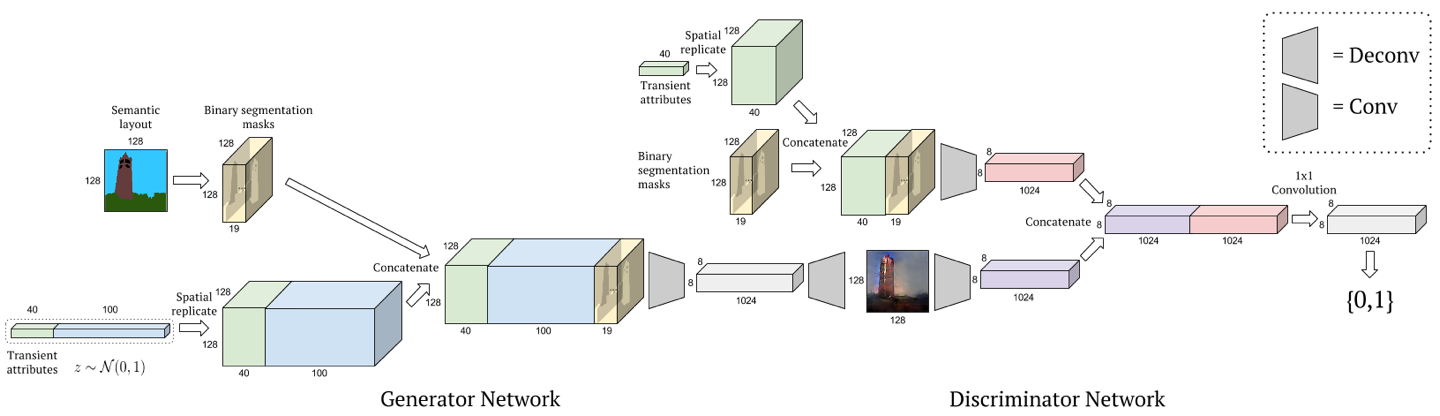}
\caption{The architectures of the generator and discriminator networks in our AL-CGAN model.}
\label{fig:model}
\end{figure*}

\myparagraph{Our Work.} Our work differs from others in the following way. We propose a novel attribute and layout conditioned GAN architecture and use it as an automatic outdoor scene editing model that learns to generate images and their edited versions both from scratch. 

\section{Model Architecture}
\label{sec:method}
In this section, we first present the main idea of generative adversarial networks (GANs) which we base our model on and then we present the details of our attribute and layout conditioned GAN (namely AL-CGAN) model.

\subsection{Preliminary: Generative Adversarial Nets}
\label{sec:GAN}
Generative adversarial networks~\cite{GPMXWDOCB14} (GANs) propose a generative model $G$ and a discriminative model $D$, which compete in a two-player min-max game. Realized as multilayer perceptrons, the discriminator model tries to accurately distinguish real images from synthetized ones while the generator tries to fool the discriminator by synthesizing images resembling real ones. Within this min-max game, the generator and the discriminator can be trained jointly by solving the following optimization problem:
\begin{eqnarray}
\min_G \max_D V(D,G) & = & E_{x \sim p_{data}(x)}[\log D(x)] + \\\nonumber
&& E_{x \sim p_{z}(z)} [\log \left(1-D(G(z))\right)]
\end{eqnarray}
where $x$ is a natural image drawn from the true data distribution $p_{data}(x)$ and $z$ is a latent random vector sampled from a uniform distribution. It is shown in~\cite{GPMXWDOCB14} that with enough number of training images and after sufficient number of epochs (i.e., if both $G$ and $D$ have enough capacity), the distribution $p_G$ converges to $p_{data}$. That is, from a random vector $z$, the generative model $G$ can synthesize an image $G(z)$ that looks like an image that is sampled from $p_{data}$. 

The conditional GAN~\cite{DCSF15, RMC16} (CGAN) is an extension of the GAN where it is augmented with some side information. Given a vector $c$ as side information, the generator $G(z,c)$ tries to synthesize a realistic image under the control of $c$. Similarly, the CGAN model allows the output of the discriminative model $D(x,c)$ to be controlled by the context vector $c$. Considering additional side information such as class labels~\cite{RMC16}, image captions~\cite{RAYLSL16}, bounding boxes and object keypoints~\cite{RAMSSL16} allows the CGAN model to generate higher quality images.

\begin{table*}[!t]
\centering
\label{tab:architecture}
\resizebox{!}{0.06\textwidth}{
\begin{tabular}{|lccccccccc|}
\hline
AL-CGAN (G) & \emph{conv1} & \emph{conv2} & \emph{conv3} & \emph{conv4} & \emph{conv5} & \emph{deconv1} & \emph{deconv2} & \emph{deconv3} & \emph{deconv4} \\ \hline \hline
Input size & 128 & 128 & 64 & 32 & 16 & 8 & 16 & 32 & 64\\ 
Kernel number & 159 & 128 & 256 & 512 & 1024 & 512 & 256 & 128 & 3\\ 
Kernel size & 5 & 5 & 5 & 5 & 5 & 5 & 5 & 5 & 5\\ 
Stride & 1 & 2 & 2 & 2 & 2 & 2 & 2 & 2 & 2\\\hline           \end{tabular}
} 
\\ \vspace{0.1cm}
\resizebox{!}{0.06\textwidth}{
\begin{tabular}{|lccccccc|}
\hline
AL-CGAN (D) & \emph{conv1}(AL/I) & \emph{conv2}(AL/I) & \emph{conv3}(AL/I) & \emph{conv4}(AL/I) & \emph{conv5}(AL/I) & \emph{conv6} & \emph{fc} \\ \hline \hline
Input size & 128/128 & 128/128 & 64/64 & 32/32 & 16/16 & 8 & - \\
Kernel number & 59/3 & 128/128 & 256/256 & 512/512 & 1024/1024 & 2048 & 1024$\times$8$\times$8 \\
Kernel size & 5/5 & 5/5 & 5/5 & 5/5 & 5/5 & 1 & - \\
Stride & 1/1 & 2/2 & 2/2 & 2/2 & 2/2 & 1 & - \\
\hline
\end{tabular}
}
\caption{Network architectures of AL-CGAN. Top: Generator network, Bottom: Discriminator network.  \emph{conv}, \emph{deconv} and \emph{fc} mean convolutional, deconvolutional layers, and fully-connected layers, respectively. Stride value 2 indicates 2$\times$ resolution. Within the Siamese architecture of the discriminator, AL and I denote attribute-layout and image networks.}
\label{tab:architecture}
\end{table*}

\subsection{AL-CGANs}
\label{subsec:ascgan}
We propose a novel CGAN architecture comprising of deconvolution and convolution layers which learn the layout and the content of the scene using ground truth semantic layouts and transient attributes. We term our model as Attribute-Layout Conditioned Generative Adversarial Net (AL-CGAN) and illustrate our architecture in \autoref{fig:model}. Formally, the generative and discriminator networks are denoted as $G: \mathbb{R}^Z \times \mathbb{R}^S\times \mathbb{R}^A\rightarrow \mathbb{R}^M$ and $D:  \mathbb{R}^S\times \mathbb{R}^A\rightarrow \{0, 1\}$  respectively, where the noise vector is $Z$-dim, the semantic layout is $S$-dim, the transient attribute vector is $A$-dim and the image is $M$-dim. We formulate our AL-CGAN as follows:
\begin{align}
& \min_G \max_D V(D,G)  = E_D + E_G \text{ where} \\ \nonumber
& E_D = E_{x,s,a \sim p_{data}(x,s,a)}[\log D(x,s,a)] \text { and} \\ \nonumber
& E_G = E_{x \sim p_{z}(z); s,a \sim p_{data}(s,a)} [\log \left(1-D(G(z,s,a))\right)]
\end{align}
We consider a 9-layer model for the generator module of AL-CGAN. It consists of 5 convolutional and 4 deconvolutional layers, as demonstrated in the top row of~\autoref{tab:architecture}. In particular, we draw the noise prior $z \in \mathcal{N}(0 ,1)$ and concatenate transient attributes vector and the $z$ vector. We tile the resulting vector to all spatial locations of $128\times  128$ and concatenate to $128 \times 128 \times 19$ semantic layout maps. We feed forward the resulting conditioning variables to stride $2$ convolutional layers. A synthetic image $\hat{x}$ is generated with standard deconvolutional layers. As in~\cite{RMC16}, we consider Batch Normalization~\cite{IS15} for all the layers of the generator. We use ReLU activation for all layers except the last one where we use $\tanh$ non-linearity. Our resulting generated images are $128\times128$ dimensional.
 
As illustrated in~\autoref{fig:model}, our discriminator network is a Siamese network~\cite{CHL05,JGLSS13}. While one of the networks takes the real/generated image as input, the second one processes the given attribute and the spatial layout maps. The responses of these networks are then integrated by using a convolutional fusion strategy. We give the details of the discriminator network in the bottom row of~\autoref{tab:architecture}. It is a 7-layer network with 6 convolutional layers (the 6th convolutional layer is for fusion) and 1 fully connected layer. In particular, in the discriminator network, we perform similar \mbox{stride-$2$} convolutional layers to layout maps to obtain conditioned maps. In the network, we obtain feature maps from the image and concatenate feature maps and condition maps to feed them to fully connected decision layer. Following~\cite{RMC16}, in all layers of the discriminator we use Batch Normalization~\cite{IS15} and LeakyReLU~\cite{MHN13,XWCL15} activation. 

\subsection{Training Details}
\label{sec:training}
We use a setting similar to the one in~\cite{RMC16}. All models were trained with mini-batch stochastic gradient descent (SGD) with a mini-batch size of $64$. Parameters were initialized from a zero-centered Normal distribution with standard deviation of $0.02$. We used the Adam optimizer\cite{kingma2014adam} with the learning rate value of $2\times 10^{-4}$ and the momentum value of $0.5$. We trained our models for $400$ epochs on a NVIDIA TITAN X GPU, which lasted about $3$ days. Our implementation is based on the Theano implementation of DCGAN\cite{RMC16} model.

\section{Experiments}
\label{sec:experiments}

In this section, we present our results on generating outdoor scenes conditioned on semantic layouts and transient attributes. We train our AL-GAN model on the union of two datasets, ADE20K~\cite{ZZPDBT16} and Transient Attributes~\cite{LRTQH14}, and we perform a set of experiments to assess the capacity of our model to generate diverse and realistic images.

\subsection{Datasets and Data Preprocessing}
Details of the datasets used to train our model are as follows. ADE20K~\cite{ZZPDBT16} dataset includes $22,210$ images from a diverse set of indoor and outdoor scenes. Each image has dense annotations of the background and each individual object. In this work, we only use a subset of the outdoor scene images from ADE20K dataset, as described in detail below. Transient Attributes~\cite{LRTQH14} dataset contains $8,571$ outdoor scene images captured by $101$ web-cams. In each webcam, there are perfectly aligned 60-120 images, which exhibit severe appearance changes due to variations in atmospheric conditions caused by weather, time of day and season. Each image in Transient Attributes dataset is hand-annotated with 40 transient scene attributes which encode perceived properties describing intra-scene variations, \eg{sunrise/sunset, cloudy, foggy, autumn, winter}. 

\begin{figure*}[!t]
\centering
\includegraphics[width=\linewidth]{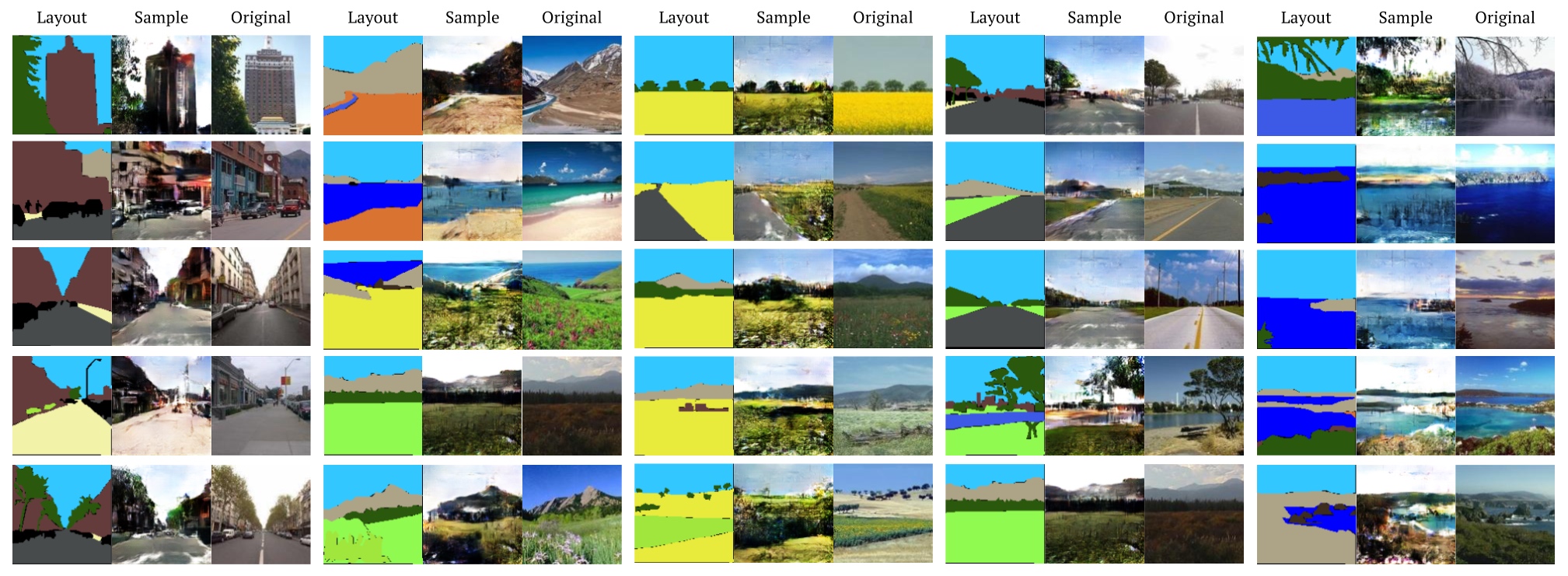}
\caption{Semantic layout conditioned outdoor scene generation using our AL-GAN. The input layouts are collected on images from SIFTflow~\cite{LYT11} and LMSun~\cite{TL13} datasets, hence they are previously unseen. The transient scene attributes are fixed to ``clear sunny day'' vector throughout the experiment.}
\label{fig:layoutconditioned}
\end{figure*}

As a pre-processing step, we first select a set of semantic labels which are commonly observed in the outdoor scene images from ADE20K and Transient Attributes datasets. These $18$ hand-picked labels are `sky', `building', `grass', `tree', `mountain', `rock', `road', `field', `ground', `earth', `sea', `water', `plant', `roof', `city', `village', `cityscape' and `hill'. 

For ADE20K, we simply employ the provided dense annotations and merge semantically similar object labels to one of our pre-defined object and background categories. For instance, the `skyscraper', `tower', `house' and `building' annotations in ADE20K are all mapped to the `building' label in our category list. For the remaining class labels, e.g. `car', `airplane', `person', etc.  we use an additional class to denote those background pixels. By this process, we have selected $9,201$ outdoor images from ADE20K with at least $70\%$ of the pixels annotated with one of our $18$ semantic labels. Note that ADE20K images do not include any transient scene attributes. Hence, we automatically predict the attributes of ADE20K images using the model in~\cite{BZGWJ16}. We empirically observed that the predicted attributes are fairly accurate, so we utilize them for training purposes. 

Since the images in Transient Attributes dataset do not have semantic layout annotations, we manually collect them using the LabelMe~\cite{LabelMe} annotation tool. As the images in each webcam are aligned, this is fairly easy. We randomly select a single image from each webcam and we only annotate that image by considering the pre-defined object and background categories. We then use the same semantic layout for all the other images from the same webcam. Note that if annotations of small scene elements such as pedestrians, cars, clouds etc., were provided, this might lead to improved results. However we avoid such dense annotations, and leave this for future work. In this way, for both datasets, we obtain $19$ non-overlapping binary layout maps with the last map denoting the unlabeled pixels. Finally, each image is resized such that the height of the output image is 128 pixels and then we take a center crop of $128\times 128$ pixels.

\subsection{Generating Realistic Outdoor Scenes}
In the following, we first present outdoor scenes generated by our AL-CGAN model using different semantic layouts. Then, we show that the model has the ability to manifest a large degree of control over the transient scene attributes. Finally, we demonstrate that it is also capable of hallucinating how the scene will look like when new scene elements are incrementally added.

\myparagraph{Effect of Varying Semantic Layouts.} Here, we synthesize novel outdoor images and observe the effect of varying semantic layouts while fixing the attribute condition vector to ``clear sunny day''. The various scene layouts that we demonstrate on~\autoref{fig:layoutconditioned} depict various scene types, e.g. urban, mountain, forest, sea, lake, and they correspond to images from SIFTflow~\cite{LYT11} and LMSun~\cite{TL13} datasets to insure that neither the original images nor the semantic layouts are previously observed while training. This way, we avoid reporting memorized scenes. The results show sharp object boundaries.  We observe no blurring effects on the boundaries of tower and skyscrapers, the ground is clearly separated from the sky and even the leafs of the trees are drawn. Similarly, realistic color distribution has been kept in our generated images, Green is the dominating color for trees and grass, while sky and sea get different shades of blue. On the other hand, buildings are colored with brown-red while roads are mostly gray. We also observe a clear scene layout in our generated images. For instance, the horizon is clearly depicted while the perspective effects such as vanishing points on the urban scenes are observed. Another observation is that our model is able to learn other scene effects such as reflection on the water, e.g. the top right-most sample. 

\begin{figure*}[!t]
\centering
\includegraphics[width=\linewidth]{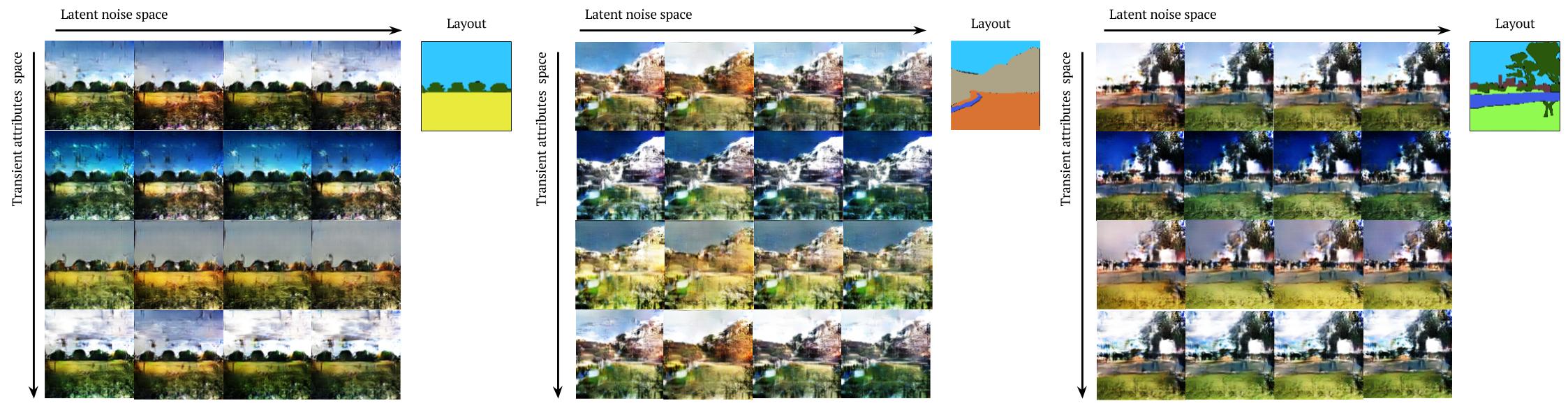}
\caption{AL-CGAN samples generated from the same semantic layout, e.g. given in the middle, by modulating the noise vector, i.e $z$. Rather than copying the previously seen scenes, our model is able to generate diverse samples.}
\label{fig:diversity}
\end{figure*}

\begin{figure*}[t]
\centering
\includegraphics[width=\linewidth]{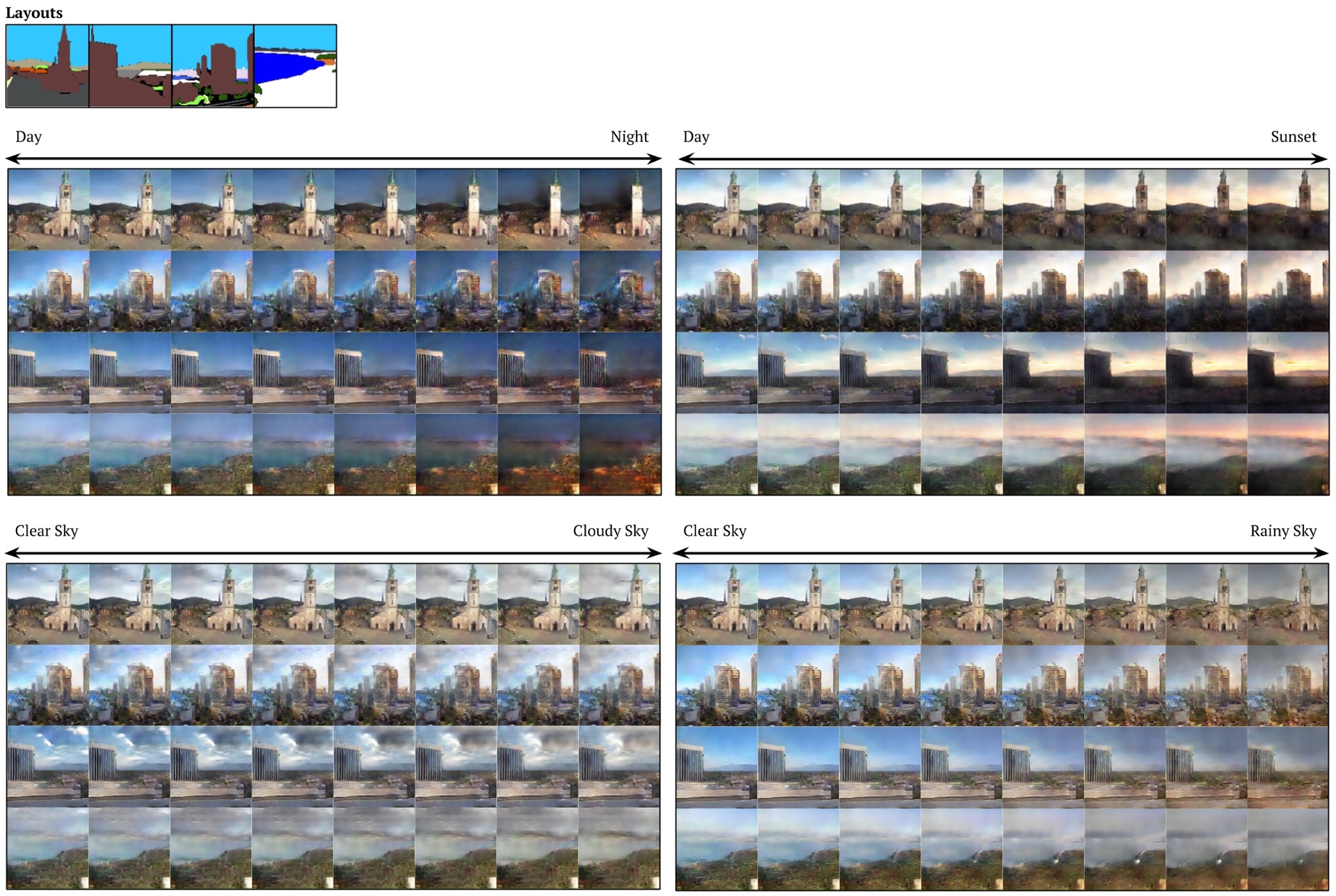}
\caption{Increasing night, sunset, cloud and rain attributes. AL-CGAN Model is trained with 9201 ADE20K images and fine tuned with images from Transient Attribute dataset (We provide more results in supplementary).}
\label{fig:attcond}
\end{figure*}

\myparagraph{Effect of Varying Transient Attributes.}
Along with generating realistic images of outdoor scenes, our second goal is to generate images of the same scene with respect to different scene conditions, e.g. transition between a sunny day and rainy day with the same scene components. As the generation outcomes is controlled by two different conditioning variables, transient attributes ($a$) and spatial layout ($s$), and the latent variable $z$, here we carry out experiments by fixing the spatial layout and then analyzing how each remaining variable controls the generation process. In Figure~\ref{fig:diversity}, we provide the generated samples for three different spatial layout maps, by varying the noise, i.e. $z$, and by varying the attributes, i.e. sunny, dark, rainy, cloudy in this example. Note that, here, neither layouts nor the corresponding images have been seen during training. By varying the noise, i.e. observe the change in horizontal direction, we generate diverse examples with varying visual aspects of the scene components, e.g. sky, color of the grass etc. to name a few. On the other hand, by varying the attributes, i.e. observe the change in vertical direction, the generated samples reflect the semantic meaning of the attribute, e.g. night attribute makes the sky darker, rain makes it gray and the density of clouds increase for the cloudy scene.

\begin{figure*}[!t]
\centering
\includegraphics[width=0.7\linewidth]{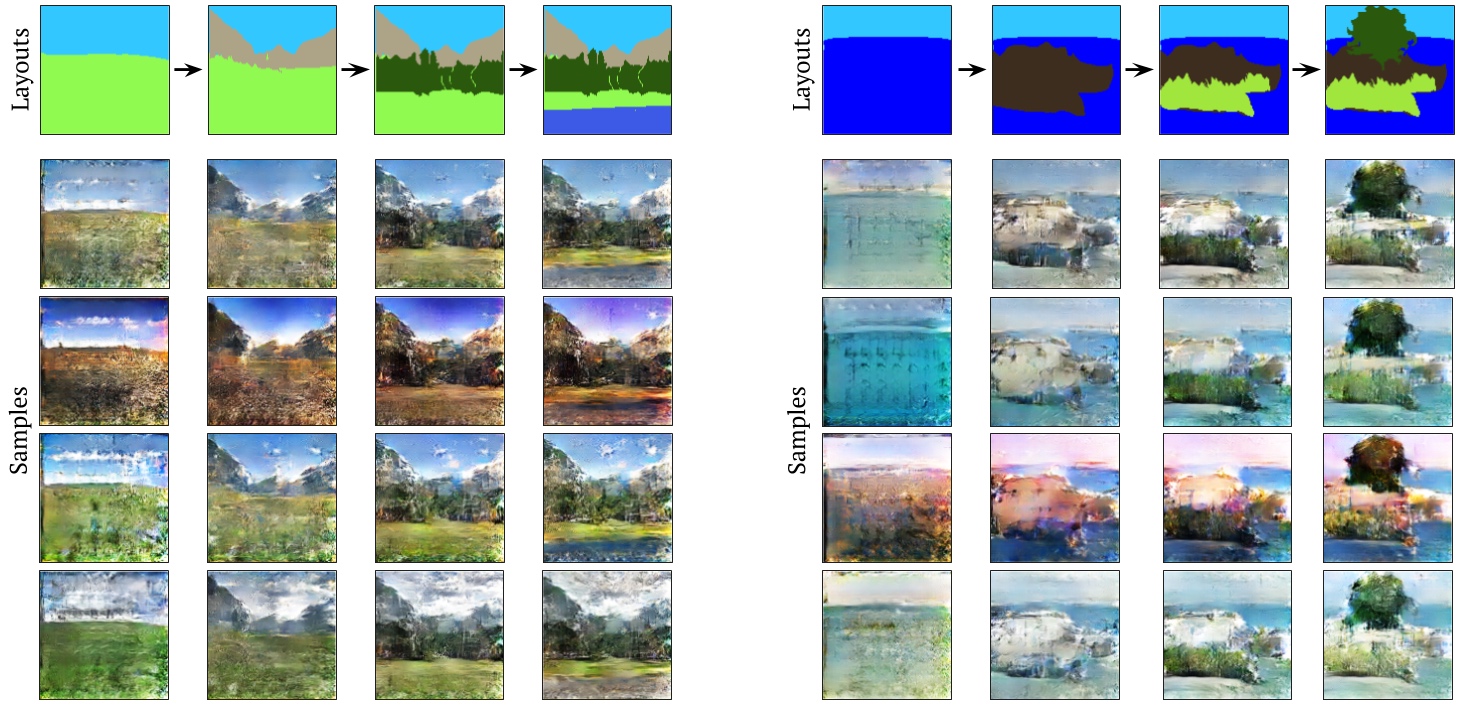}
\caption{Gradually adding details to the generated images. We employ a coarse spatial layout map to generate an image from scratch, and then keep adding new scene elements to the map to refine the generated images.}
\label{fig:adding_new_segments}
\end{figure*}

Alternatively, we evaluate our model on previously seen semantic layouts by generating images with varying attribute strength to achieve the transition between different scene conditions. As it can be seen from the results in~\autoref{fig:attcond}, our model is not only able to generate close to photo-realistic images of the scene, but it also is able to imagine how the same scene would look like at night, at sunset, with a cloudy or rainy weather. Transient Attributes dataset contains example images of this scene at night, at sunset, with clouds and under the rain, however the content of the scene itself changes with moving objects, shadows etc. Our model is able to ignore such local changes in the scene and generate a photo-realistic interpretation of transient attributes. Note that with the increasing night attribute, the buildings get illuminated whereas increasing sunset attribute darkens the buildings. On the other hand, clouds do not change the global appearance of the scene, however as expected only the relevant portion of the sky gets modified. Finally, for the rain attribute the entire scene gradually assumes a grayish tone. These results may demonstrate that our AL-CGAN model learns interesting and relevant internal representations of outdoor scenes. 

\begin{figure*}[!t] 
\centering
\includegraphics[width=0.7\linewidth]{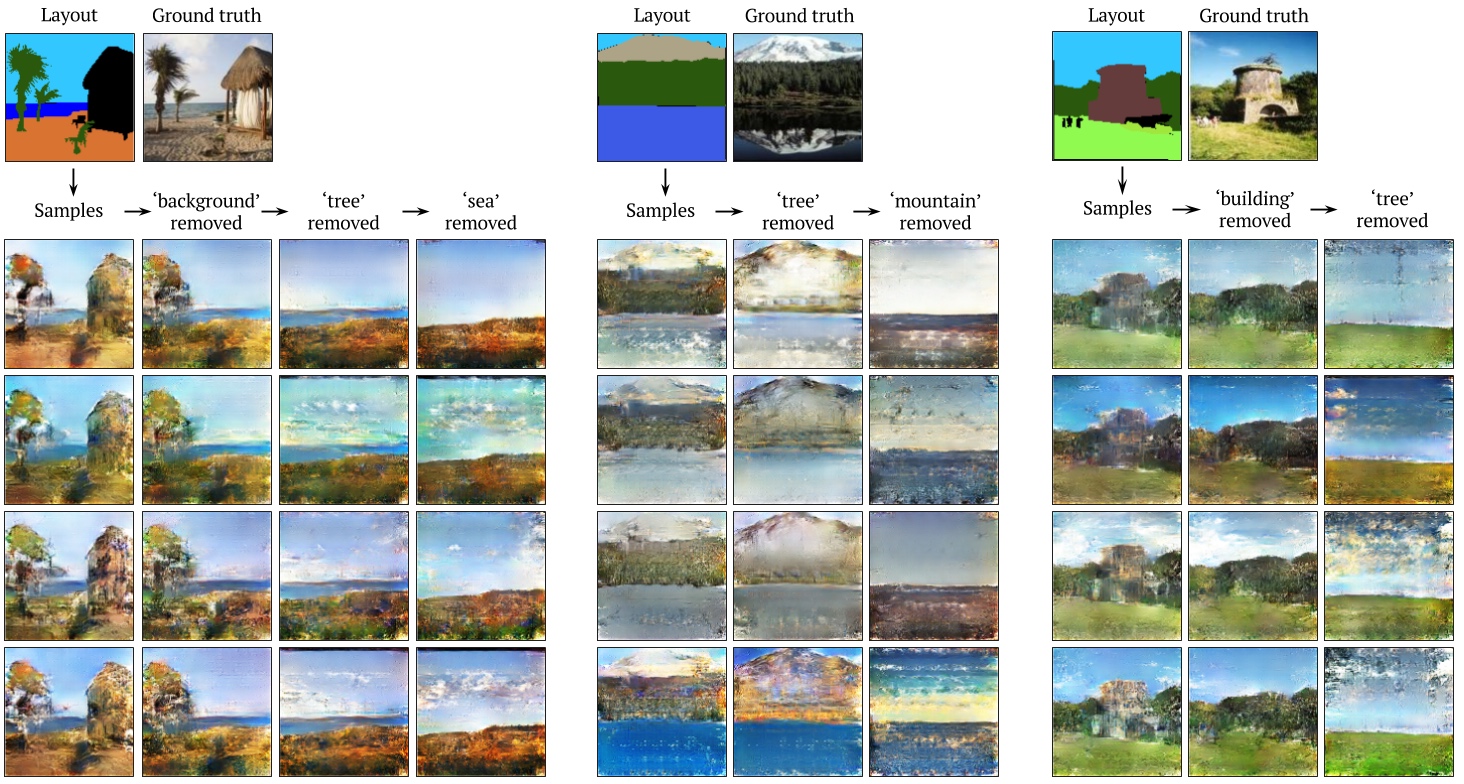}
\caption{Simplifying the generated images by erasing certain scene elements. We generate an outdoor image from scratch by using a detailed layout map, and then we keep simplifying the image by gradually erasing certain scene elements.}
\vspace{-3mm}
\label{fig:object_removal}
\end{figure*}

\myparagraph{Incrementally Adding/Deleting Scene Elements.} 
In this section, we explore one potential application of our model to generate images in an incremental manner. We begin with coarse spatial layouts which contain two large segments from different classes, e.g. sky and grass. We then gradually add new scene elements, e.g. mountain, tree, lake. At each step, the model generates a new natural image that best satisfies the given semantic layout and the provided scene attributes. In Figure~\ref{fig:adding_new_segments}, we present the outcomes of two such experiments  where we consider different semantic categories. For the image given on the left, we start with a simple scene containing only sky and grass regions. Consequently, we insert mountains and forest into the background and a lake to the front in an iterative way. As it can be seen from these results, adding each new scene element results in a more detailed image. On the generated images from the last step, we even observe the reflection of the mountains and the forest over the lake. Similar observations can be made for the second sequence of layouts. The coarse samples from the first step contain only sea and sky regions but the next round of images become more and more detailed with the inclusion of a rocky island, grasslands on the island, and finally adding a lone tree to the corresponding semantic layouts. Note that scene guidance in the form of semantic layouts leads to more detailed and thus realistic scenes. Our conclusion from these results is that for methods such as GAN to generate realistic scenes, the type, the location and the shape of the conditioning variables are important. In fact, we suspect that this approach closely resembles human thought process in imagining and painting novel scenes as Bob Ross, the famous painter whom we borrowed the quote in the beginning of our paper, repeatedly describes in his instructive painting classes. 

\begin{figure*}[!t]
\centering
\includegraphics[width=\linewidth]{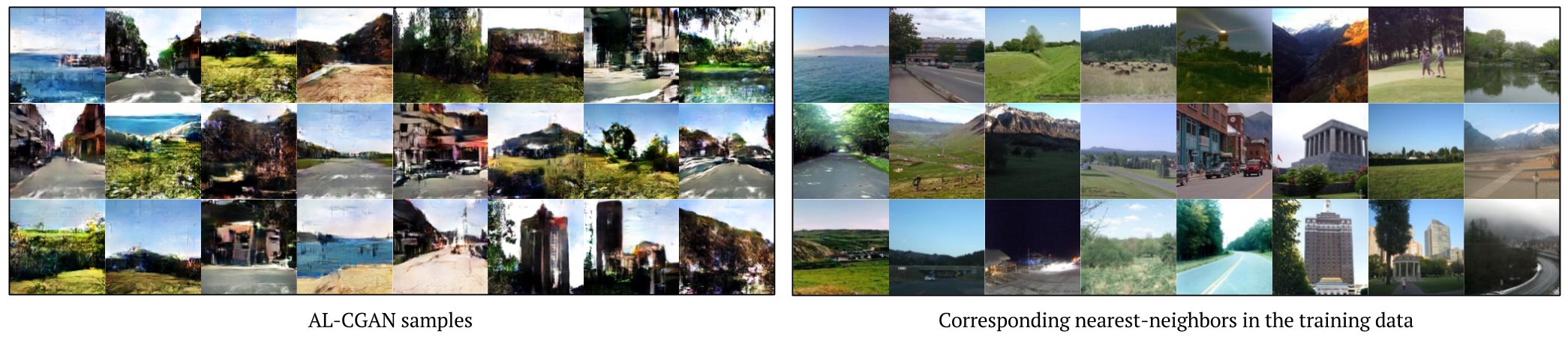}
\caption{The nearest training images for some samples from our AL-GAN model.}
\vspace{-3mm}
\label{fig:overfit}
\end{figure*}

We can also condition the generation process on coarser scene layouts while keeping the noise vector and the transient attributes fixed. Specifically, we start with a previously seen scene layout, and we keep erasing a specific scene element from the semantic map to produce simplified versions of the original scene. Figure~\ref{fig:object_removal} presents  samples we generate in this set of experiments. Our AL-CGAN model can produce convincing but less detailed images than those from the previous steps of the generation process.

\myparagraph{Searching for Nearest Training Images.} 
As a sanity check to inspect whether the generated samples are indeed diverse, we perform an additional set of experiments where we find the nearest training images (according to $l_1$ distance in raw image space) for some sample images generated by our AL-CGAN model. We present our results in Figure~\ref{fig:overfit}. We observe that the images and their nearest neighbors are in most cases semantically related and have similar scene structures. On the other hand, for some of the synthesized images, the scene categories of the corresponding nearest neighbors are different, or in some cases, even if their classes are the same, the  scene elements in the pair of images are quite different. For instance, for the leftmost image in the middle row, while the generated image depicts an urban scene, its nearest neighbor is an image of a rural scene. Similarly, the sea image in the top-most left corner resembles its nearest neighbor, e.g. being also a sea image, however, the urban scene image right below it is completely different from the forest image that happens to be its nearest neighbor. These results indicate that in most cases our novel AL-CGAN architecture does not memorize the scenes but generate images of novel scenes from scratch.

\subsection{Comparing with Other GAN Architectures}
\begin{figure*}[!t]
\centering
\includegraphics[width=1\linewidth]{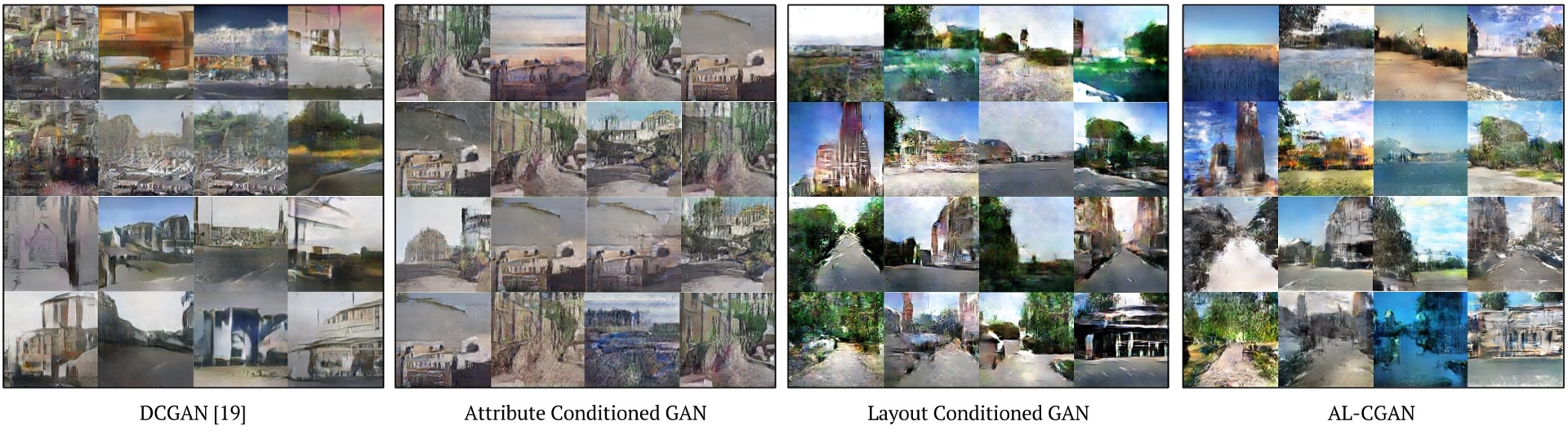} \\
\caption{Qualitative comparison of AL-CGAN samples against  DCGAN~\cite{RMC16}, i.e. scene label conditioning, and our model with three ablations. AL-CGAN is our model with both attribute and scene layout conditioning, A-CGAN is our model with only attribute conditioning and L-GAN is our model with scene layout conditioning. We observe that generated images get sharper and more realistic when more conditioning variables are added.}
\vspace{-2mm}
\label{fig:comparison}
\end{figure*}

In this section, we compare the scene label conditioned GAN baseline~\cite{RMC16} with our AL-CGAN model which generates an image under the condition of attributes and spatial layout. We also provide an ablation study on the outcomes of two versions of our AL-CGAN model, i.e. only-attribute-conditioned A-CGAN and only-layout-conditioned L-CGAN models. Finally, we generate images using our full AL-CGAN model that includes both attribute and semantic layout conditioning components. In Figure~\ref{fig:comparison}, we show sample results of these network models, which are all trained on the same data containing images from ADE20K and Transient Attributes datasets. 

Our first observation is that the scene label conditioned GAN~\cite{RMC16}, already generates images of plausible scenes however the color distribution does not show much variation and the details of the scene elements are not present. On the other hand, the attribute conditioned version of our AL-CGAN model generates images of a similar nature. Although the images clearly show outdoor scenes, we observe repeated objects and the color distribution is monotonous. We observe that already with our spatial layouts conditioned architecture, i.e. L-CGAN, the objects are formed with more clarity especially on the boundaries. Furthermore, the images are sharper, more clear and semantically more meaningful. Finally, our complete AL-CGAN model that uses both attribute and scene layout conditioning leads to more diverse scenes with more details, more realistic color distribution and even sharper object boundaries. These results suggest that providing additional side information in the form of conditioning variables is helpful for learning to generate better natural-looking images.

\section{Conclusion}
\label{sec:conc}
In this work, we proposed a novel conditioned GAN (CGAN) architecture that is able to generate realistic outdoor scenes under the guidance of semantic layouts to specify where to draw background and objects using the specified transient attributes such as day-night and/or sunny-foggy that give directions on how the global appearance characteristics should be encoded. Our novel novel deep conditional generative adversarial network architecture called the AL-CGAN model, employs spatially replicated transient scene attributes and pixel-based semantic labels as condition vectors. We showed that by varying the semantic layouts, we can control the objects drawn in the image within the specified semantic boundaries. By varying transient attributes, we showed that our AL-CGAN generates scenes with various conditions, i.e. sunny, cloudy etc and it allows a smooth transition between transient attributes. We further demonstrated that our AL-CGAN model can generate more detailed images by gradually adding new scene elements. Our ablation study showed that every component of our framework is necessary for higher quality images. As a future work, we plan to extend our model so that it generates realistic images of natural language descriptions along with using semantic layouts. 

\section*{Acknowledgments}
We gratefully acknowledge the support of NVIDIA Corporation with the donation of the Tesla K40 GPU used in this study.
{\small
\bibliographystyle{ieee}
\bibliography{egbib}
}

\clearpage
\renewcommand\thesection{S.\arabic{section}}
\setcounter{section}{0}

\begin{table*}[!t]
\centering
\begin{tabular}{c}
\textbf{\Large Additional Results}
\end{tabular}
\end{table*}

\begin{figure*}[!t] 
\begin{center}
\includegraphics[width=0.47\linewidth]{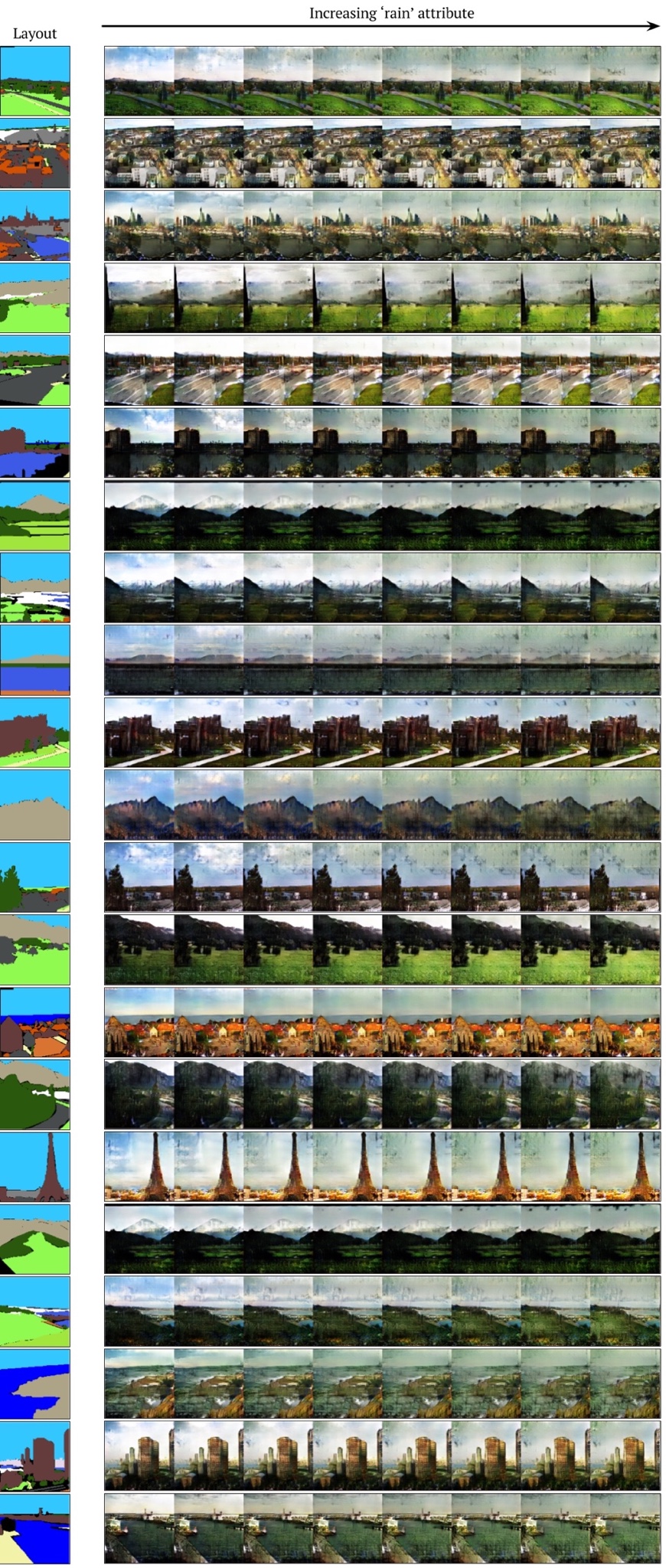}$\quad$
\includegraphics[width=0.47\linewidth]{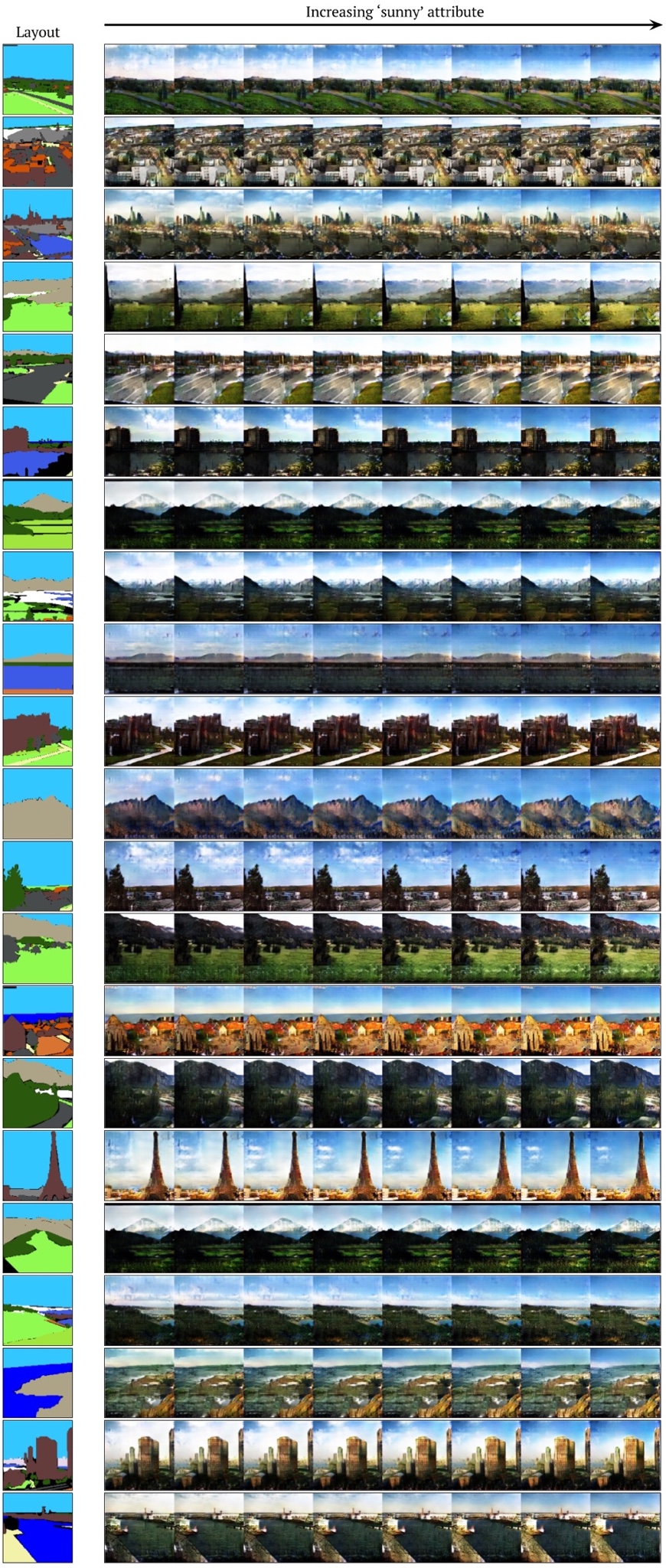}
\caption{AL-CGAN results on adjusting `rain' and `sunny' attributes. The model is trained with images from ADE20K and Transient Attributes dataset, and samples are generated using layouts seen during the training.}
\end{center}
\end{figure*}

\begin{figure*}[!t] 
\begin{center}
\includegraphics[width=0.47\linewidth]{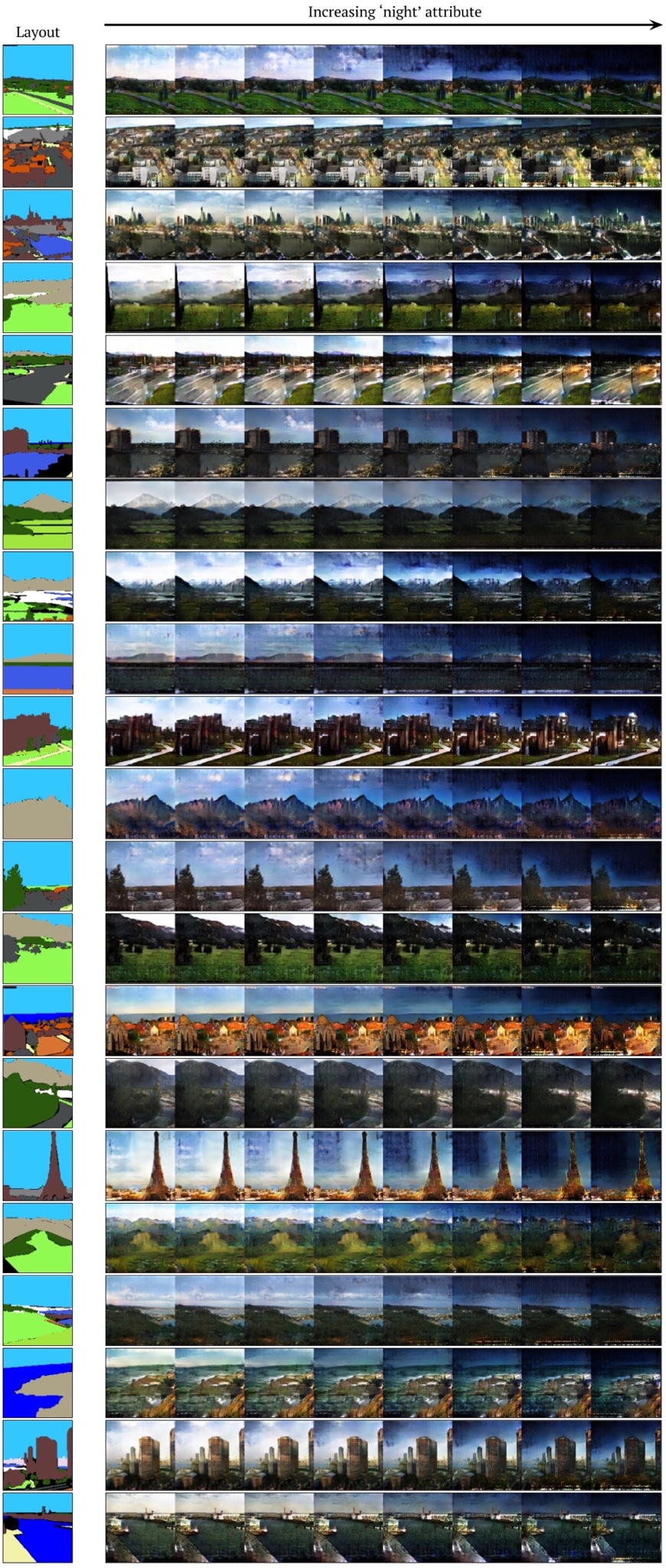}$\quad$
\includegraphics[width=0.47\linewidth]{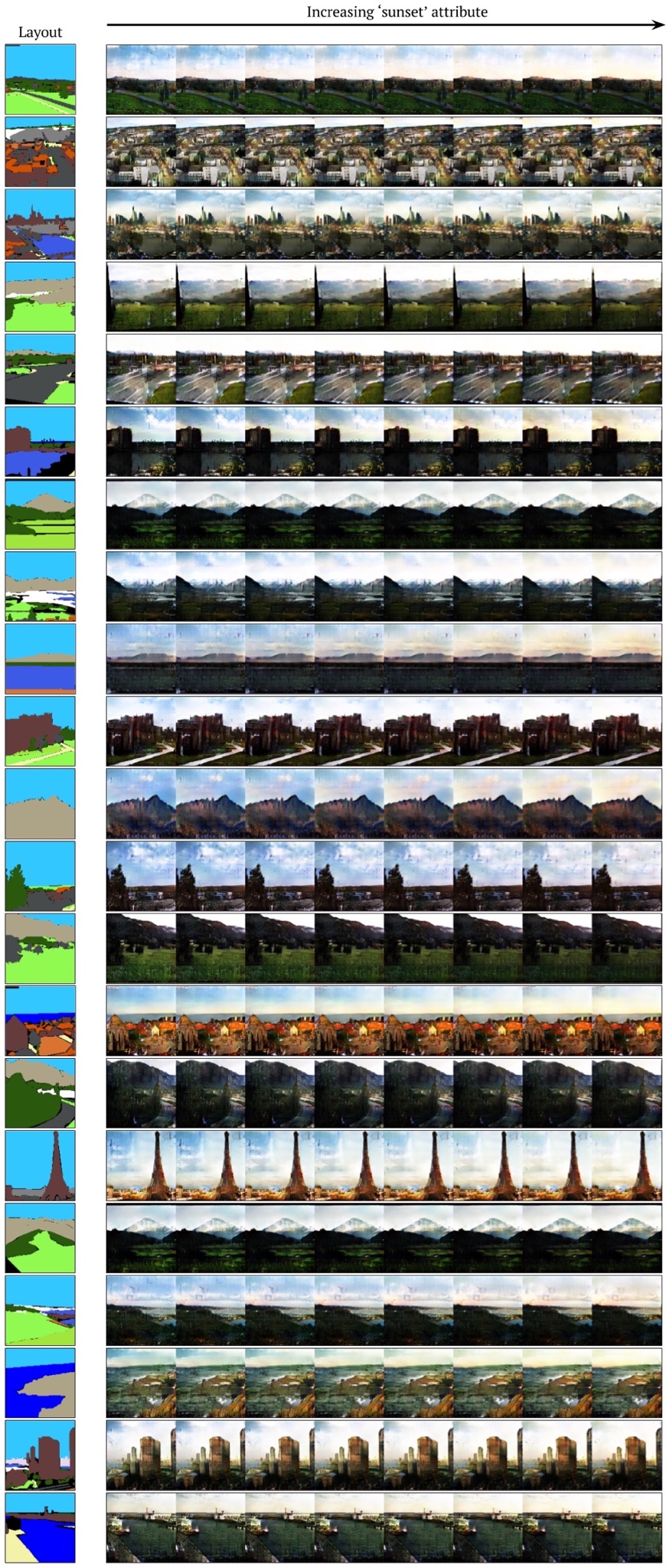}
\caption{AL-CGAN results on adjusting `night' and `sunset' attributes. The model is trained with images from ADE20K and Transient Attributes dataset, and samples are generated using layouts seen during the training.}
\end{center}
\end{figure*}

\begin{figure*}[!t] 
\begin{center}
\includegraphics[width=0.47\linewidth]{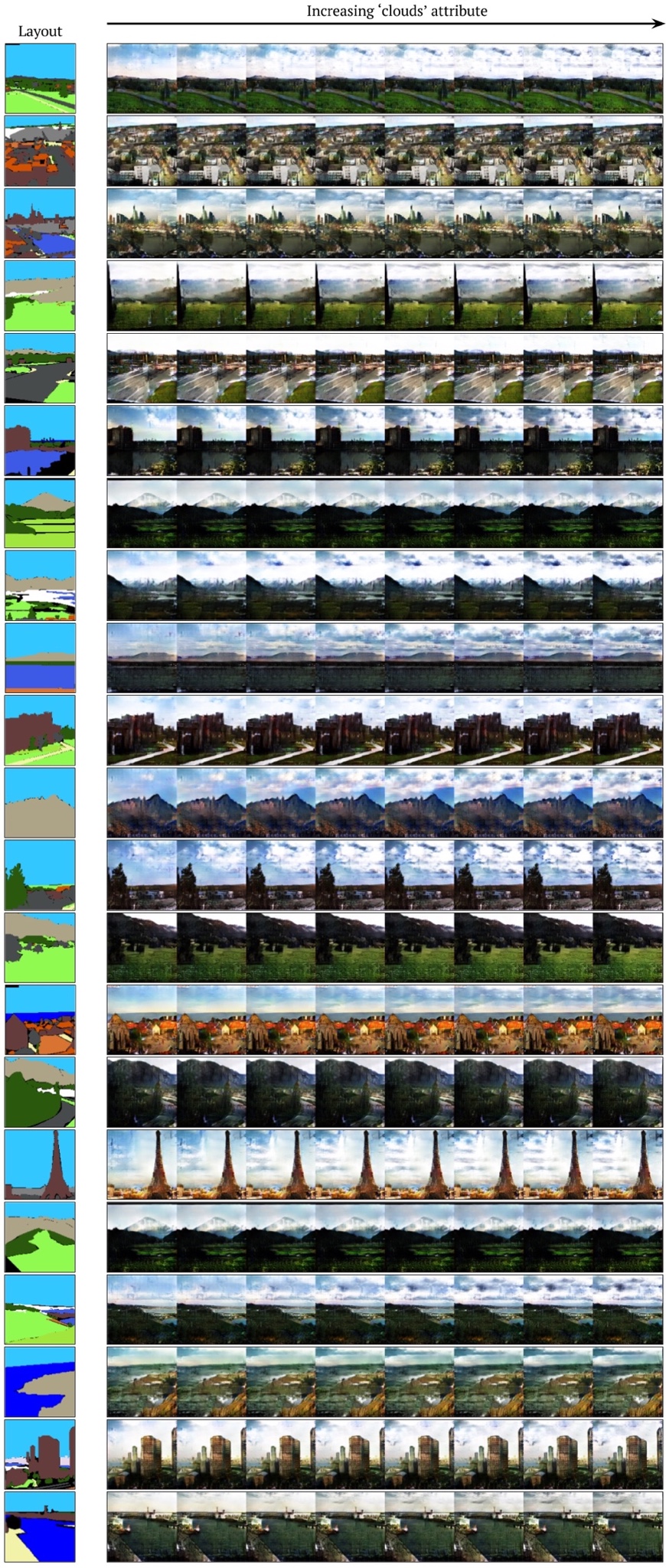}$\quad$
\includegraphics[width=0.47\linewidth]{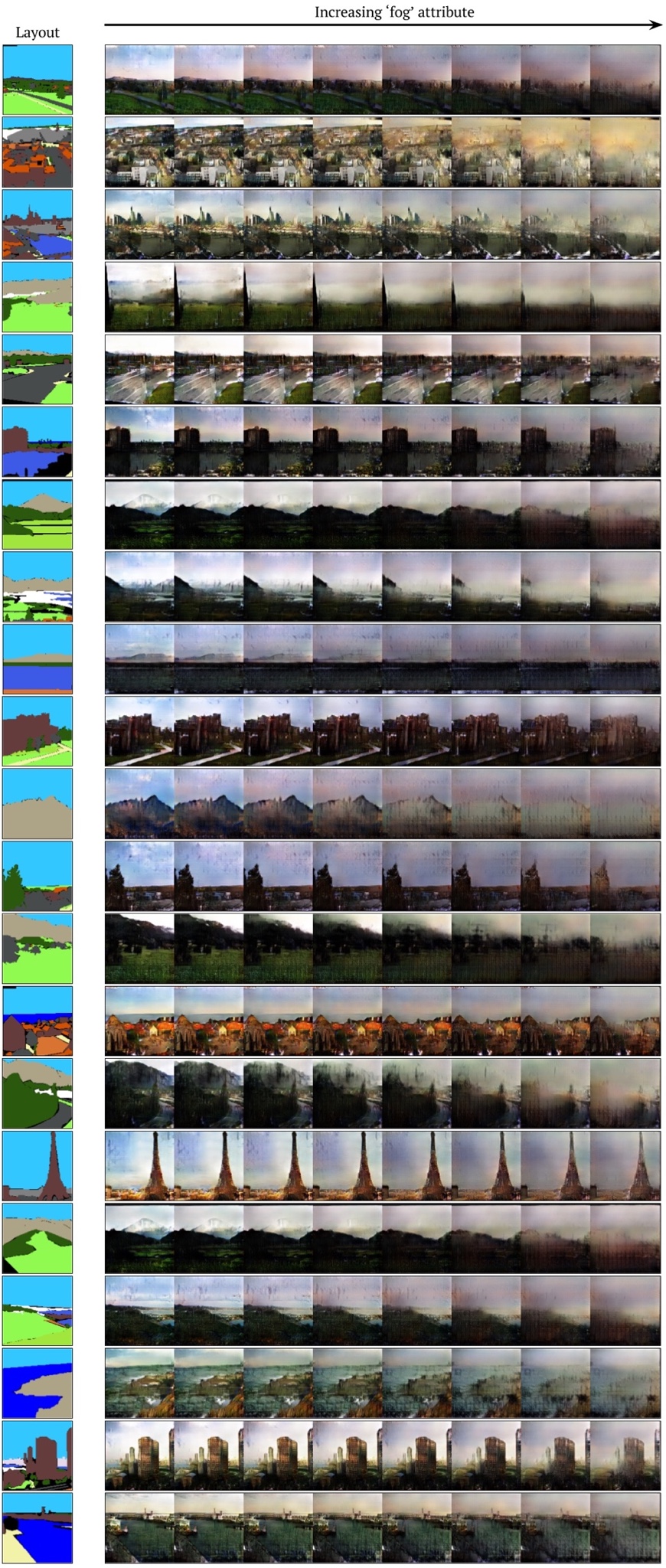}
\caption{AL	-CGAN results on adjusting `clouds' and `fog' attributes. The model is trained with images from ADE20K and Transient Attributes dataset, and samples are generated using layouts seen during the training.}
\end{center}
\end{figure*}

\begin{figure*}[!ht] 
\centering
\includegraphics[width=\linewidth]{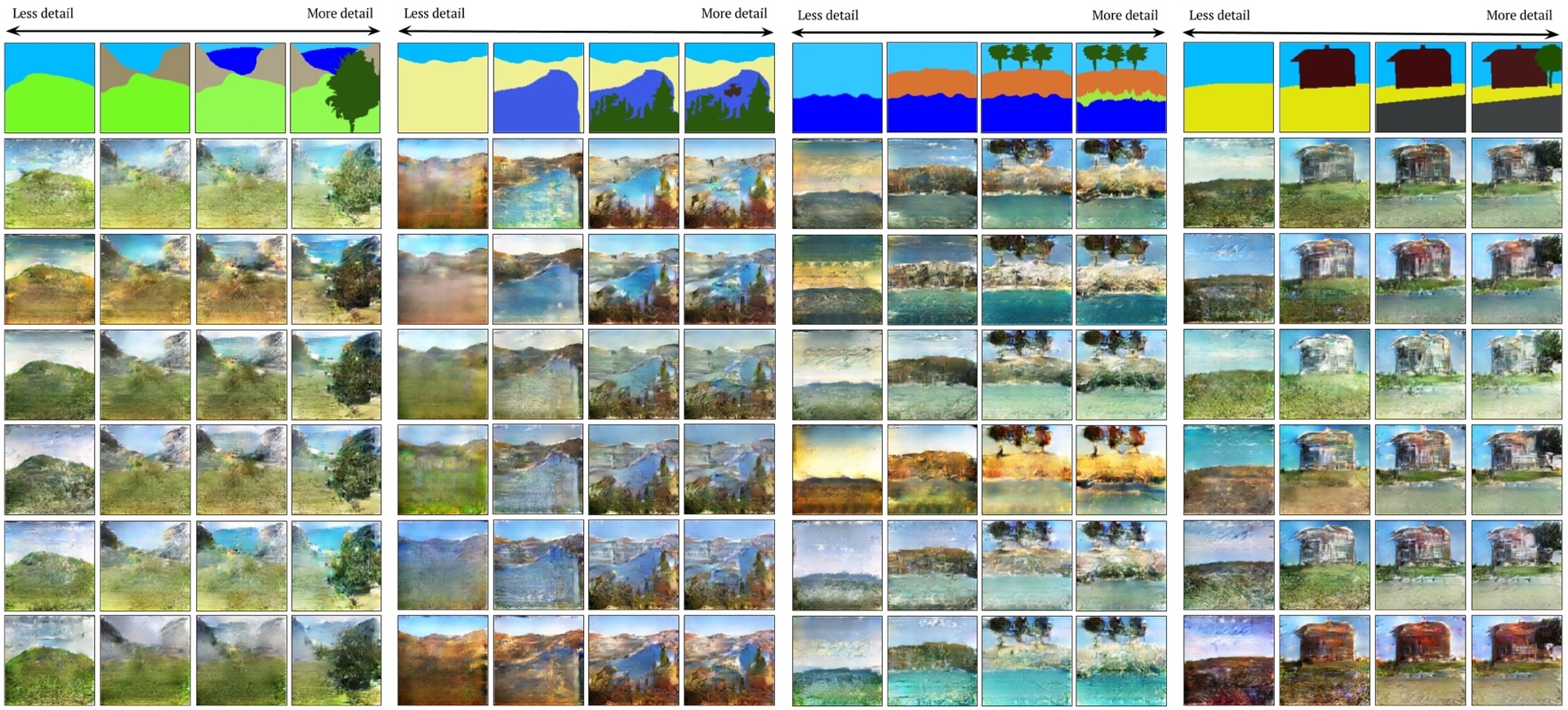}
\caption{Incrementally adding news scene elements. All semantic layouts are novel, provided by the user and have not been seen before.}
\vspace{-3mm}
\label{fig:scene_creation}
\end{figure*}

\begin{figure*}[!t] 
\includegraphics[width=\linewidth]{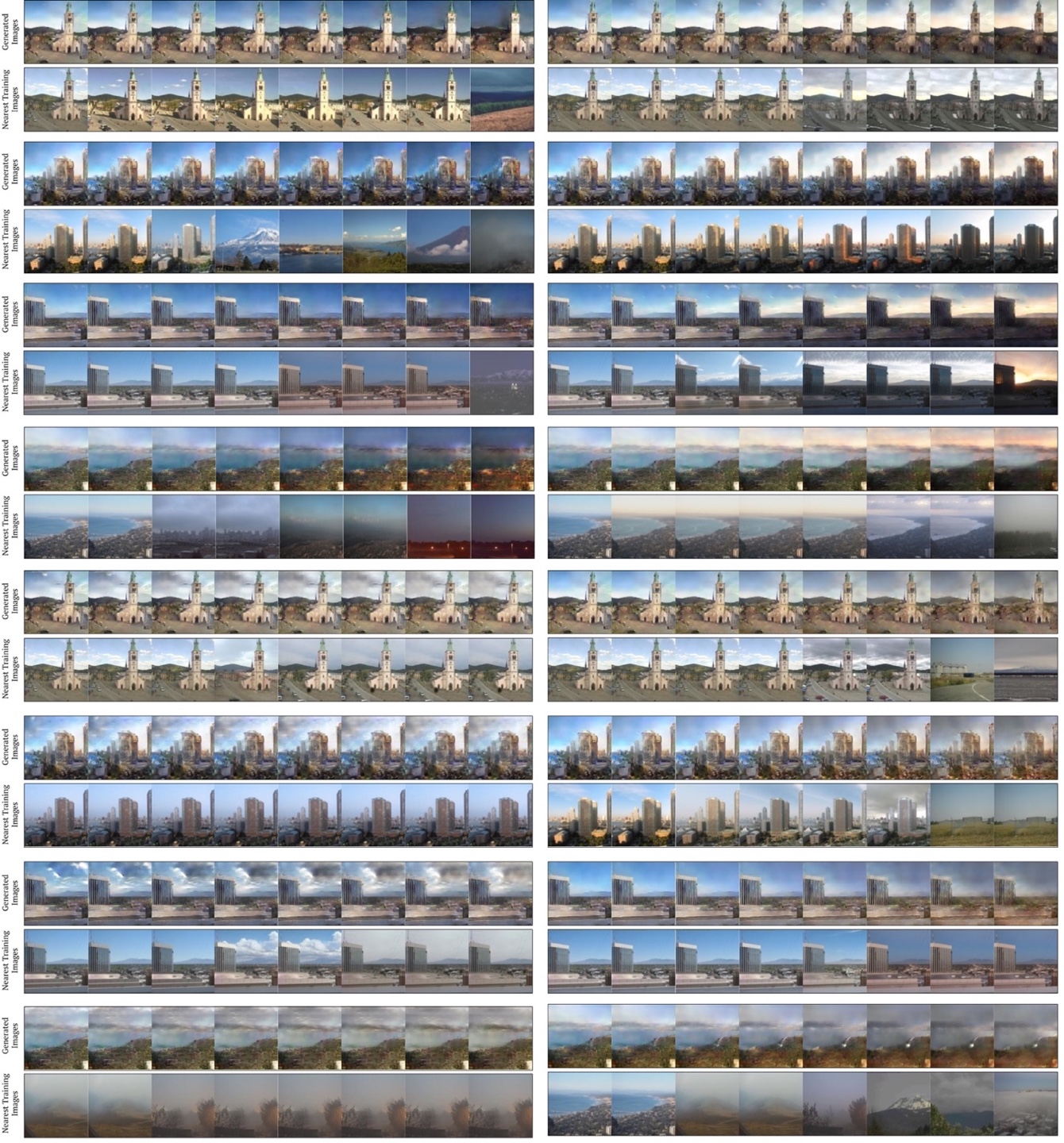}
\caption{Nearest training images for the synthesized images lying on manifolds of different attributes. While the interpolation carried along the learned image manifold gives smooth changes in the scene characteristics, the sequences obtained by the corresponding nearest training images lack such kind of reasonable transformations. For some of the generated images, the corresponding nearest images are even from different scenes.}
\vspace{-3mm}
\label{fig:nearest}
\end{figure*}

\end{document}